%% file: emnlp-ijcnlp-2019.tex
\documentclass[11pt,a4paper]{article}
\usepackage[hyperref]{emnlp-ijcnlp-2019}
\usepackage{times}
\usepackage{latexsym}

\usepackage{url}
\usepackage{flushend}

\input{sections/usepackage.tex}

\aclfinalcopy 

\title{Learning to Discriminate Perturbations for\\Blocking Adversarial Attacks in Text Classification}

\author{Yichao Zhou$^*$, Jyun-Yu Jiang$^*$, Kai-Wei Chang and Wei Wang\\
  Computer Science Department \\
  University of California, Los Angeles\\
  {\tt \{yz, jyunyu, kwchang, weiwang\}@cs.ucla.edu}}

\date{}

\begin{document}
\maketitle

\blfootnote{$^*$Equal contribution. Listing order is random.}
\begin{abstract}
\input{sections/s0-abstract.tex}
\end{abstract}

\input{sections/s1-intro.tex}

\input{sections/s2-relatedwork.tex}

\input{sections/s3-methodology.tex}
\input{sections/s4-exp.tex}
\input{sections/s5-conclusions.tex}

\section*{Acknowledgment}

We would like to thank the anonymous reviewers for their helpful comments. The work was supported by NSF DGE-1829071 and NSF IIS-1760523.

\bibliography{emnlp-ijcnlp-2019}
\bibliographystyle{acl_natbib}

\end{document}

%% file: sections/usepackage.tex
\usepackage{multirow}
\usepackage{graphicx}
\usepackage{bbm}
\usepackage{bm}
\usepackage{amsmath}
\usepackage{amsfonts} 
\usepackage{amssymb}
\usepackage{enumitem}
\usepackage{subcaption}
\usepackage[normalem]{ulem}

\usepackage[ruled,vlined,linesnumbered]{algorithm2e}

\newcommand{\mR}{\mathbb{R}}

\newcommand{\argmax}[1]{\underset{#1}{\operatorname{argmax}}}

\usepackage[symbol]{footmisc}

\newcommand\blfootnote[1]{%
  \begingroup
  \renewcommand\thefootnote{}\footnote{#1}%
  \addtocounter{footnote}{-1}%
  \endgroup
}

%% file: sections/s0-abstract.tex
Adversarial attacks against machine learning models have  threatened various real-world applications such as spam filtering and sentiment analysis.
In this paper, we propose a novel framework, learning to \underline{dis}criminate \underline{p}erturbations (DISP), to identify and adjust malicious perturbations, thereby blocking adversarial attacks for text classification models.
To identify adversarial attacks, a perturbation discriminator validates how likely a token in the text is perturbed and provides a set of potential perturbations.
For each potential perturbation, 
an embedding estimator learns to restore the embedding of the original word based on the context and a replacement token is chosen  based on approximate $k$NN search.
DISP can block adversarial attacks for any NLP model without modifying the model structure or training procedure.
Extensive experiments on two benchmark datasets demonstrate that DISP significantly outperforms baseline methods in blocking adversarial attacks for text classification.
In addition, in-depth analysis shows the robustness of DISP across different situations.

%% file: sections/s1-intro.tex
\section{Introduction}
\label{section:intro}

Deep learning techniques~\cite{goodfellow2016deep} have achieved enormous success in many fields, such as computer vision and NLP.
However, complex deep learning models are often sensitive and vulnerable to a tiny modification.
In other words, malicious attackers can destroy the models by adding a few inconspicuous perturbations into input data, such as masking images with unrecognizable filters and making low-key modifications for texts.
Therefore, developing techniques to equip models against adversarial attacks becomes a prominent research problem.

Existing studies on adversarial attacks can be classified into two groups, generation of adversarial examples and defense against adversarial attacks~\cite{yuan2019adversarial}.
In the field of NLP, most of the existing studies focus on the former.
For example, \newcite{ebrahimi2017hotflip,alzantot2018generating} replace a word with synonyms or similar words while \newcite{gao2018black,liang2017deep,ebrahimi2017hotflip} conduct character-level manipulations to fool the models.
Moreover, it is not straightforward to adapt existing approaches for blocking adversarial attacks, such as data augmentation~\cite{krizhevsky2012imagenet,ribeiro2018semantically,ren2019generating} and adversarial training~\cite{goodfellow2015explaining,iyyer2018adversarial,marzinotto2019robust,cheng2019evaluating,zhu2019retrieval}, to NLP applications. 
Hence, the defense against adversarial attacks in NLP remains a challenging and unsolved problem.

Recognizing and removing the inconspicuous perturbations are the core of defense against adversarial attacks.
For instance, in computer vision, denoising auto-encoders~\cite{warde2017improving, gu2015towards} are applied to remove the noises introduced by perturbations; \citet{prakash2018deflecting} manipulate the images to make the trained models more robust to the perturbations; \citet{samangouei2018defense} apply generative adversarial networks to generate perturbation-free images.
However, all of these approaches cannot straightforwardly apply to the NLP tasks for the following two reasons.
First, images consist of continuous pixels while texts are discrete tokens.
As a result, a token can be replaced with another semantically similar token that drops the performance, so perturbations with natural looks cannot be easily recognized compared to previous approaches that capture unusual differences between the intensities of neighboring pixels.
Second, sentences consist of words with an enormous vocabulary size, so it is intractable to enumerate all of the possible sentences.
Therefore, existing defense approaches in computer vision that rely on pixel intensities cannot be directly used for the NLP tasks.

After recognizing the perturbed tokens, the na\"ive way to eliminate the perturbations for blocking adversarial attacks is to remove these perturbed tokens.
However, removing words from sentences results in fractured sentences, causing the performance of NLP models to degrade.  
Therefore, it is essential to recover the removed tokens. 
Nevertheless, training a satisfactory language model requires myriad and diverse training data, which is often unavailable.
An inaccurate language model that incoherently patches missing tokens can further worsen the prediction performance.
To tackle this difficult challenge, we propose to recover the tokens from discriminated perturbations by a masked language model objective with contextualized language modeling.

In this paper, we propose \emph{Learning to \underline{Dis}criminate \underline{P}erturbations} (DISP), as a framework for blocking adversarial attacks in NLP.
More specifically, we aim to defend the model against adversarial attacks without modifying the model structure and the training procedure.
DISP consists of three components, perturbation discriminator, embedding estimator, and hierarchical navigable small world graphs.
Given a perturbed testing data, the perturbation discriminator first identifies a set of perturbed tokens.
For each perturbed token, the embedding estimator optimized with a corpus of token embeddings infers an embedding vector to represent its semantics.
Finally, we conduct an efficient $k$NN search over a hierarchical taxonomy 
to translate each of the embedding vectors into appropriate token to replace the associated perturbed word.
We summarize our contributions in the following.

\begin{itemize}[leftmargin=*]
    \item To the best of our knowledge, this paper is the first work for blocking adversarial attacks in NLP without retraining the model.
    \item We propose a novel framework, DISP, which is effective and 
    significantly outperforms other baseline methods in defense against adversarial attacks on two benchmark datasets.
    \item Comprehensive experiments have been conducted to demonstrate the improvements of DISP. In addition, we will release our implementations and the datasets to provide a testbed and facilitate future research in this direction.
\end{itemize}

%% file: sections/s2-relatedwork.tex
\section{Related Work}
\label{section:relatedwork}

Adversarial examples crafted by malicious attackers expose the vulnerability of deep neural networks when they are applied to down-streaming tasks, such as image recognition, speech processing, and text classifications~\cite{wang2019survey,goodfellow2015explaining, nguyen2015deep, moosavi2017universal}. 

For adversarial attacks, white-box attacks have full access to the target model while black-box attacks can only explore the models by observing the outputs with limited trials.  \citet{ebrahimi2017hotflip} propose a gradient-based white-box model to attack character-level classifiers via an atomic flip operation. 
Small character-level transformations, such as swap, deletion, and insertion, are applied on critical tokens identified with a scoring strategy~\cite{gao2018black} or gradient-based computation~\cite{liang2017deep}.
\citet{samanta2017towards, alzantot2018generating} replace words with semantically and syntactically similar adversarial examples. 

However, limited efforts have been done on adversarial defense in the NLP fields. Texts as discrete data are sensitive to the perturbations and cannot transplant most of the defense techniques from the image processing domain such as Gaussian denoising with autoencoders~\cite{meng2017magnet, gu2014towards}. Adversarial training is the prevailing counter-measure to build a robust model~\cite{goodfellow2015explaining,iyyer2018adversarial,marzinotto2019robust,cheng2019evaluating,zhu2019retrieval} by mixing adversarial examples with the original ones during training the model. However, these adversarial examples can be detected and deactivated by a genetic algorithm~\cite{alzantot2018generating}. This method also requires retraining, which can be time and cost consuming for large-scale models.

Spelling correction~\cite{mays1991context, islam2009real} and grammar error correction~\cite{sakaguchi2017grammatical} are useful tools which can block editorial adversarial attacks, such as swap and insertion. However, they cannot handle cases where word-level attacks that do not cause spelling and grammar errors. In our paper, we propose a general schema to block both word-level and character-level attacks. 

%% file: sections/s3-methodology.tex
\section{DISP for Blocking Adversarial Attacks}
\label{section:method}

\begin{figure*}[!t]
    \centering
    \includegraphics[width=.8\linewidth]{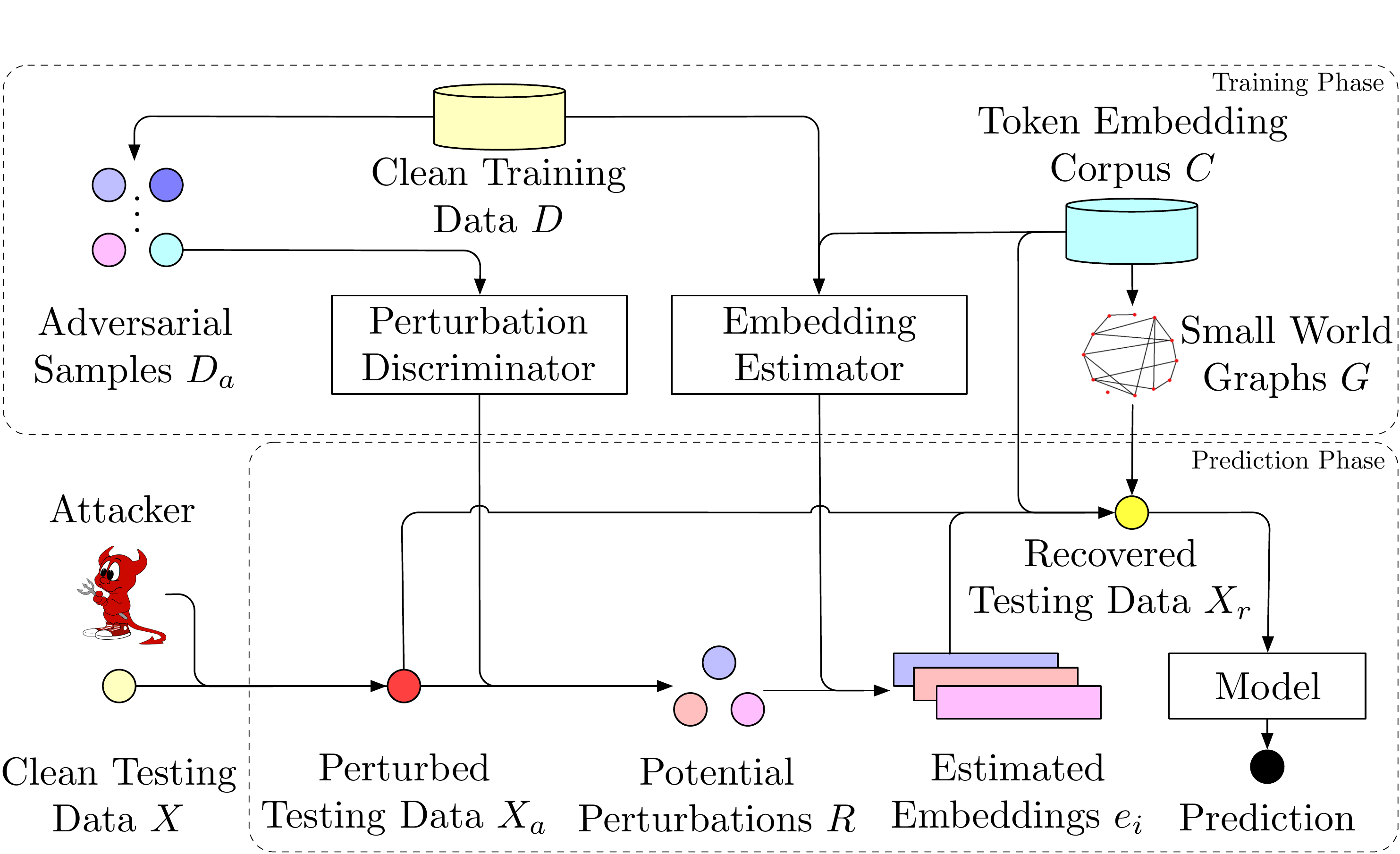}
    \caption{Schema of the proposed framework DISP.}
    \label{fig:schema}
\end{figure*}

In this section, we first formally define the goal of adversarial defense and then introduce the proposed framework DISP, learning to \underline{dis}criminate \underline{p}erturbations, for blocking adversarial attacks.

\noindent \textbf{Problem Statement.}
Given an NLP model $F(X)$, where $X = \lbrace t_1, \dots, t_N\rbrace$ is the input text of $N$ tokens while $t_i$ indicates the $i$-th token.
A malicious attacker can add a few inconspicuous perturbations into the input text and generate an adversarial example $X_a$ so that $F(X)\neq F(X_a)$ with unsatisfactory prediction performance.
For example, a perturbation can be an insertion, a deletion of a character in a token, a replacement of a token with its synonym.
In this paper, we aim to block adversarial attacks for general text classification models.
More specifically, we seek to preserve the model performances by recovering original input text and universally improve the robustness of any text classification model.

\subsection{Framework Overview}

Figure~\ref{fig:schema} illustrates the overall schema of the proposed framework.
DISP consists of three components, (1)~a perturbation discriminator, (2)~an embedding estimator, and (3)~a token embedding corpus with the corresponding small world graphs $G$.
In the training phase, DISP constructs a corpus $D$ from the original corpus for training the perturbation discriminator so that it is capable of recognizing the perturbed tokens.
The corpus of token embeddings $C$ is then applied to train the embedding estimator to recover the removed tokens after establishing the small world graphs $G$ of the embedding corpus.
In the prediction phase, for each token in testing data, the perturbation discriminator predicts if the token is perturbed.
For each potential perturbation that is potentially perturbed, the embedding estimator generates an approximate embedding vector and retrieve the token with the closest distance in the embedding space for token recovery.
Finally, the recovered testing data can be applied for prediction.
Note that the prediction model can be any NLP model.
Moreover, DISP is a general framework for blocking adversarial attacks, so the model selection for the discriminator and estimator can also be flexible.

\subsection{Perturbation Discrimination}

\begin{figure}[!t]
    \centering
    \includegraphics[width=\linewidth]{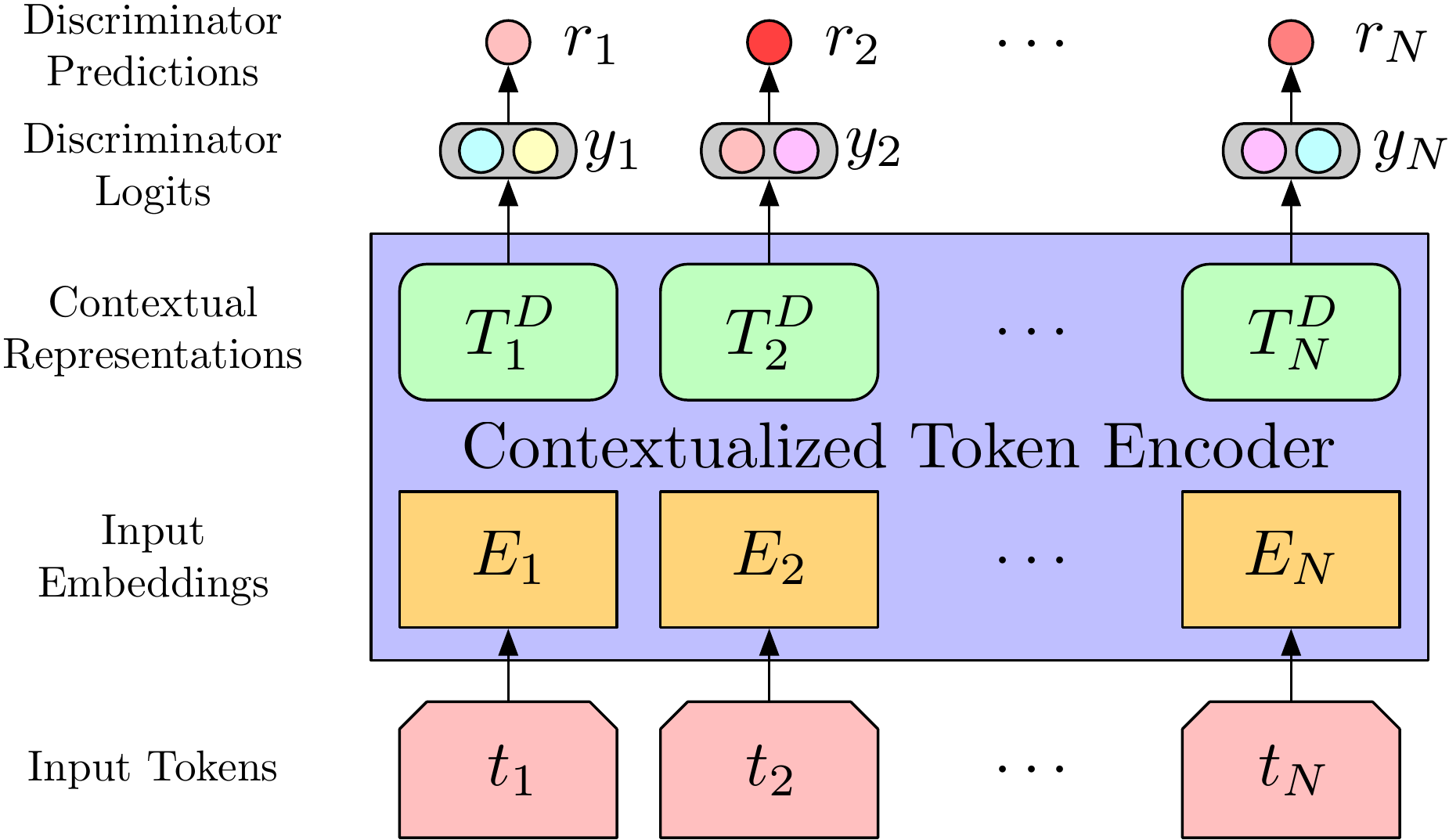}
    \caption{The illustration of the perturbation discriminator in DISP.}
    \label{fig:discriminator}
\end{figure}

\noindent \textbf{Perturbation Discriminator}.
The perturbation discriminator plays an important role to classify whether a token $t_i$ in the input $X_a$ is perturbed based on its neighboring tokens.
We adopt contextualized language modeling, such as BERT~\cite{devlin2018bert}, to derive $d$-dimension contextualized token representation $T^D_i$ for each token $t_i$ and then cascade it with a binary logistic regression classifier to predict if the token $t_i$ is perturbed or not. 
Figure~\ref{fig:discriminator} illustrates the perturbation discriminator based on a contextualized word encoder.
The discriminator classifies a token $t_i$ into two classes $\lbrace0, 1\rbrace$ with logistic regression based on the contextual representation $T^D_i$ to indicate if the token is perturbed.
More formally, for each token $t_i$, the discriminator predictions $r_i$ can then be derived as:
$$r_i = \argmax{c}~y^c_i = \argmax{c}\left(\bm{w_c}\cdot T^D_i + b_c\right), $$
where $y^c_i$ is the logit for the class $c$; $\bm{w_c}$ and $b_c$ are the weights and the bias for the class $c$.
Finally, the potential perturbations $R$ is the set of tokens with positive discriminator predictions $R=\left\lbrace t_i \mid r_i=1 \right\rbrace$.

\subsection{Efficient Token-level Recovery with Embedding Estimator}

After predicting the perturbations $R$, we need to correct these disorders to preserve the prediction performance.
One of the most intuitive approaches to recover tokens with context is to exploit language models.
However, language models require sufficient training data while the precision to exact tokens can be dispensable for rescuing prediction performance.
Moreover, over-fitting limited training data can be harmful to the prediction quality.
To resolve this problem, we assume that replacing the perturbed word with a word with similar meanings to the original word is sufficient for the downstream models to make the correct prediction. 
Based on the assumption, DISP first predicts the embeddings of the recovered tokens  for the potential perturbations with an embedding estimator based on context tokens.
The tokens can then be appropriately recovered by an efficient $k$-nearest neighbors ($k$NN) search in the embedding space of a token embedding corpus $C$.

\noindent \textbf{Embedding Estimator.}
Similar to the perturbation discriminator, any regression model can be employed as an embedding estimator based on the proposed concept.
Here we adopt the contextualized language modeling again as an example of the embedding estimator.
For each token $t_i$, the contextualized token embedding can be derived as a $d$-dimensional contextual representation vector $T^G_i$ to be features for estimating appropriate embeddings.

\begin{figure}[!t]
    \centering
    \includegraphics[width=\linewidth]{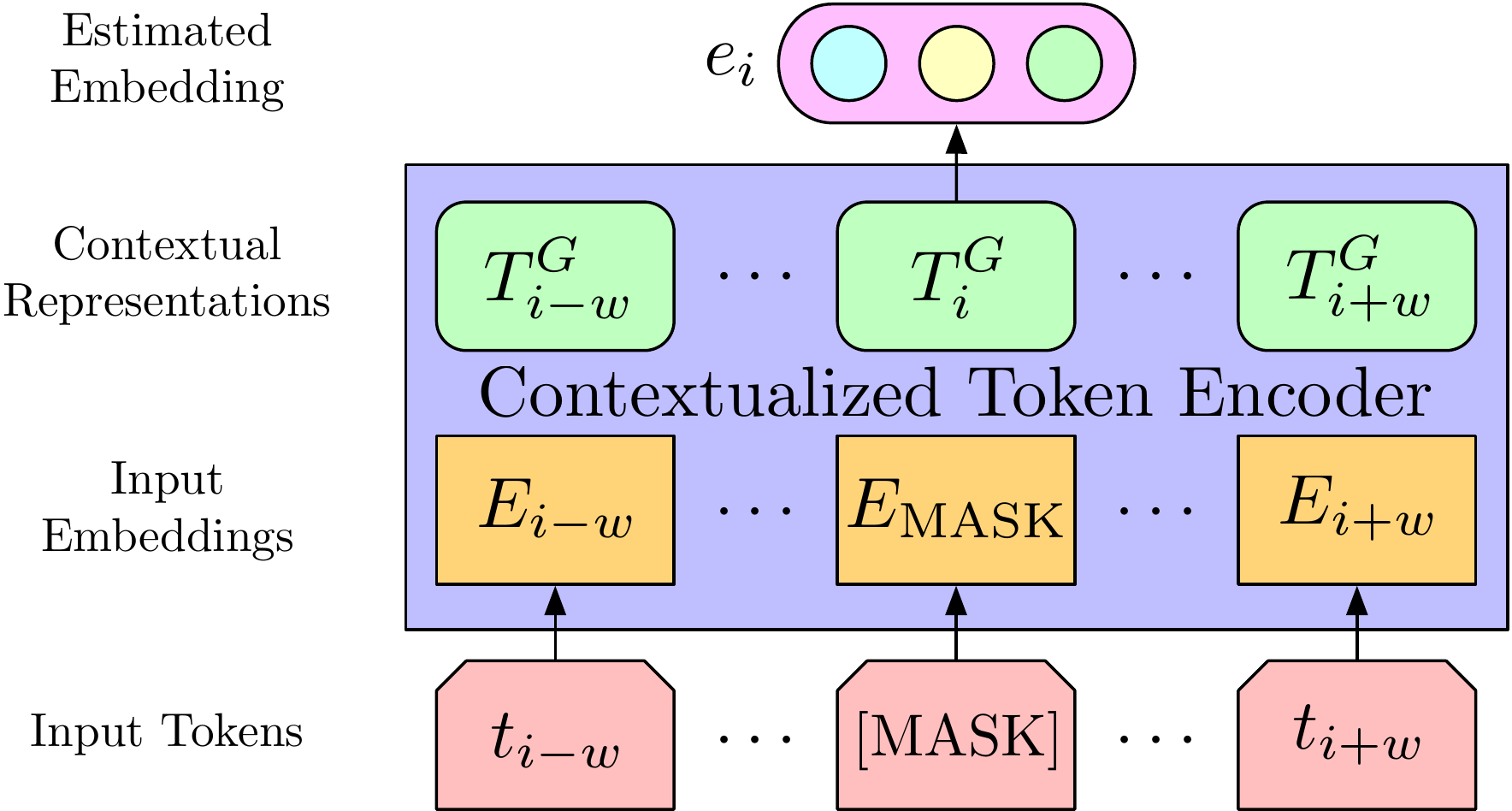}
    \caption{The illustration of the embedding estimator in DISP with a window size $2w+1$ for the token at the position $i$.}
    \label{fig:estimator}
\end{figure}
 
Figure~\ref{fig:estimator} shows the embedding estimator based on BERT. 
For each potential perturbation $t_i \in R$, $2w$ neighboring tokens are selected as the context for estimating the appropriate embedding, where $w$ decides the window size.
More precisely, a segment of tokens with a window size $2w+1$ from $t_{i-w}$ to $t_{i+w}$ is the input tokens for BERT, where $t_i$ is replaced with a \texttt{[MASK]} token as the perturbed position.
Finally, for the target $t_i$, a weight matrix $W^G\in \mR^{d\times k}$ projects the contextual representation $T^G_i$ to a $k$-dimensional estimated embedding $e_i$ as follows:
$$e_i = T^{G}_{i} W^G,$$
where the dimension size $k$ is required to be consistent with the embedding dimension in the token embedding corpus $C$.

\noindent \textbf{Efficient Token-level Recovery.}
Finally, we recover the input sentence based on the predicted recover embeddings from the embedding estimator.
Specifically, the input text $X$ needs to be recovered from the perturbed text $X_a$ by fixing token-level perturbations based on its approximate embeddings.

Given the token embedding corpus $C$, it is simple to transform an embedding to a token by finding the nearest neighbor token in the embedding space.
However, a na\"ive $k$NN search query can take $O(kn)$ time complexity, where $n$ is the number of embeddings in $C$; $k$ is the embedding dimension.
To accelerate the search process, we apply hierarchical navigable small world graphs~(SWGs)~\cite{malkov2018efficient} for fast approximate $k$NN search.
More precisely, embeddings are transformed into a hierarchical set of SWGs based on the proximity between different embeddings.
To conduct $k$NN searches, the property of degree distributions in SWGs significantly reduces the search space of each $k$NN query from $O(n)$ to $O(\log n)$ by navigating on the graphs, so a $k$NN query can be efficiently completed in $O(k\log n)$ time complexity.
Finally, the recovered text $X_r$ can be obtained by replacing the perturbations $R$ in $X_a$ as shown in Algorithm~\ref{algo:recover}.

\begin{algorithm}[!t]
    \caption{Efficient Token-level Recovery}
    \label{algo:recover} 
    \KwIn{\small Perturbed text $X_a$, potential perturbations $R$, estimated embeddings $\lbrace e_i \rbrace$, small world graphs $G$, token embedding corpus $C$.}
    \KwOut{\small Recovered text $X_r$.}
    $X_r$ = $X_a$;\\
    \For{$t_i \in R$}{
        index = QuerySmallWorldGraph($G$, $e_i$);\\
        $z$ = $C$[index].token;\\ 
        Replace $t_i$ in $X_r$ with $z$;\\
	}
  	\Return $X_r$;\\
\end{algorithm}

\subsection{Learning and Optimization}

To learn a robust discriminator, we randomly sample adversarial examples from both character-level and word-level attacks in each training epoch.
The loss function optimizes the cross-entropy between the labels and the probabilistic scores computed by the logits $y_i$ and the softmax function.

The learning process of embedding estimator is similar to masked language models.  
The major difference is that language models optimize the likelihood to generate the same original token while the embedding estimator minimizes the distance between the derived embedding and the original token embedding.
To learn the embedding estimator, a size-$(2w+1)$ sliding window is applied to enumerate $(2w+1)$-gram training data for approximating embeddings with context tokens.
For optimization, the embedding estimator is learned to minimize the mean square error~(MSE) from the inferred embeddings to the original token embeddings.


To take advantage of hierarchical navigable SWGs for an efficient recovery,
although a pre-process to construct SWGs $G$ is required, the pre-process can be fast. The established SWGs can also be serialized in advance.
More precisely, the time complexity is $O(kn\log n)$ for one-time construction of reusable SWGs, where $n$ is the number of embeddings in the embedding corpus $C$.

%% file: sections/s4-exp.tex
\section{Experiments}
\label{section:exp}

In this section, we conduct extensive experiments to evaluate the performance of DISP in improving model robustness.
\input{tables/datasets.tex}
\input{tables/attacks.tex}

\input{tables/discriminator_exp.tex}

\subsection{Experimental Settings}

\noindent \textbf{Experimental Datasets.} Experiments are conducted on two benchmark datasets: (1) 
Stanford Sentiment Treebank Binary (SST-2)~\cite{socher2013recursive} and (2) Internet Movie Database (IMDb)~\cite{maas2011learning}. 
SST-2 and IMDb are both sentiment classification datasets which involve binary labels annotating sentiment of sentences in movie reviews.  
Detailed statistics of two datasets are listed in Table~\ref{tab:dataset}.

\noindent \textbf{Attack Generation.}
We consider  three types of character-level attacks and two types of word-level attacks.
The character-level attacks consist of {\textit{insertion}}, {\textit{deletion}}, and {\textit{swap}}.
\textit{Insertion} and \textit{deletion} attacks inject and remove a character, respectively,  while a \textit{swap} attack flips two adjacent characters.
The word-level attacks include \textit{random} and \textit{embed}.
A \textit{random} attack randomly samples a word to replace the target word while a \textit{embed} attack replaces the word with a word among the top-10 nearest words in the embedding space.
The examples of each attack type are illustrated in Table~\ref{tab:attacks}. 
To obtain strong adversarial attack samples,w we consider to leverage oracle to identify the perturbations that cause prediction changes. Specifically, for each test sample we construct 50 adversarial examples by perturbing the test data. We sample one example in which model prediction changes after perturbing. 
If none of them can change the prediction, the sample with the least confidence is selected.

\input{tables/mainexp.tex}

\noindent \textbf{Base Model and Baselines.}
We consider {BERT}~\cite{devlin2018bert} as the base model as it achieves strong performance in these benchmarks.  
To evaluate the performance of DISP, we consider the following baseline methods: {(1) Adversarial Data Augmentation~(ADA)} samples adversarial examples to increase the diversity of training data;  {(2) Adversarial Training~(AT)} samples different adversarial examples in each training epoch; {(3) Spelling Correction~(SC)} is used as a baseline for discriminating perturbations and blocking character-level attacks. 
Note that  ADA and AT require to re-train BERT with the augmented training data, while
DISP and SC modify the input text and then exploit the original model for prediction.
SC is also the only baseline for evaluating discriminator performance.
In addition, we also try to ensemble DISP and SC (DISP+SC) by conducting DISP on the spelling corrected input.

\noindent \textbf{Evaluation Metrics.}
We evaluate the performance of the perturbation discriminator by precision, recall and F1 scores, and evaluate the overall end-to-end performance by classification accuracy that the models recover.  

\noindent \textbf{Implementation Details.} The model is implemented in PyTorch~\cite{paszke2017automatic}. 
We set the initial learning and dropout parameter to be $2\times 10^{-5}$ and 0.1. We use crawl-300d-2M word embeddings from \textit{fastText}~\cite{mikolov2018advances} to search similar words. The dimensions of word embedding $k$ and contextual representation $d$ are set as 300 and 768. $w$ is set as 2.
We follow $\text{BERT}_{\text{BASE}}$ \cite{devlin2018bert} to set the numbers of layers (i.e., Transformer blocks) and self-attention heads as 12. 

\subsection{Experimental Results}

\noindent \textbf{Performance on identifying perpetuated tokens.}
Table~\ref{tab:disc_exp} shows the performance of DISP and SC in discriminating perturbations.
Compared to SC, DISP has an absolute improvement by 35\% and 46\% on SST-2 and IMDb in terms of F1-score, respectively.
It also proves that the context information is essential when discriminating the perturbations.
An interesting observation is that SC has high recall but low precision scores for character-level attacks because it is eager to correct misspellings while most of its corrections are not perturbations.
Conversely, DISP has more balances of recall and precision scores since it is optimized to discriminate the perturbed tokens.
For the word-level attacks, SC shows similar low performance on both \textit{random} and \textit{embed} attacks while DISP behaves much better.
Moreover, DISP works better on the \textit{random} attack because the embeddings of the original tokens tend to have noticeably greater Euclidean distances to randomly-picked tokens than the distances to other tokens.

\noindent \textbf{Defense Performance.} 
Table~\ref{tab:main_exp} reports the accuracy scores of all methods with different types of adversarial attacks on two datasets.
Compared to the baseline BERT model, all of the methods alleviate the performance drops.
All methods perform better on blocking character-level attacks than word-level attacks because word-level attacks eliminate more information.
For the baselines, consistent with Table~\ref{tab:disc_exp}, SC performs the best for character-level attacks and the worst for word-level attacks.
In contrast, ADA and AT are comparably more stable across different types of attacks.
The differences between performance for character- and word-level attacks are less obvious in IMDb because documents in IMDb tend to be longer with more contexts to support the models.
DISP works well to block all types of attacks. Compared with the best baseline models, DISP significantly improves the classification accuracy by 2.51\% and 5.10\% for SST-2 and IMDb, respectively. By ensembling SC and DISP, DISP+SC achieves better performance for blocking all types of attacks.
However, the improvements are not consistent in IMDb.  In particular, SC performs worse with lower discrimination accuracy and over-correcting the documents.
In addition, DISP has a stable defense performance across different types of attacks on IMDb because richer context information in the documents benefits token recovery.

\begin{figure*}[!t]
    \centering
    \begin{subfigure}[t]{.33\linewidth}
        \includegraphics[width=\linewidth]{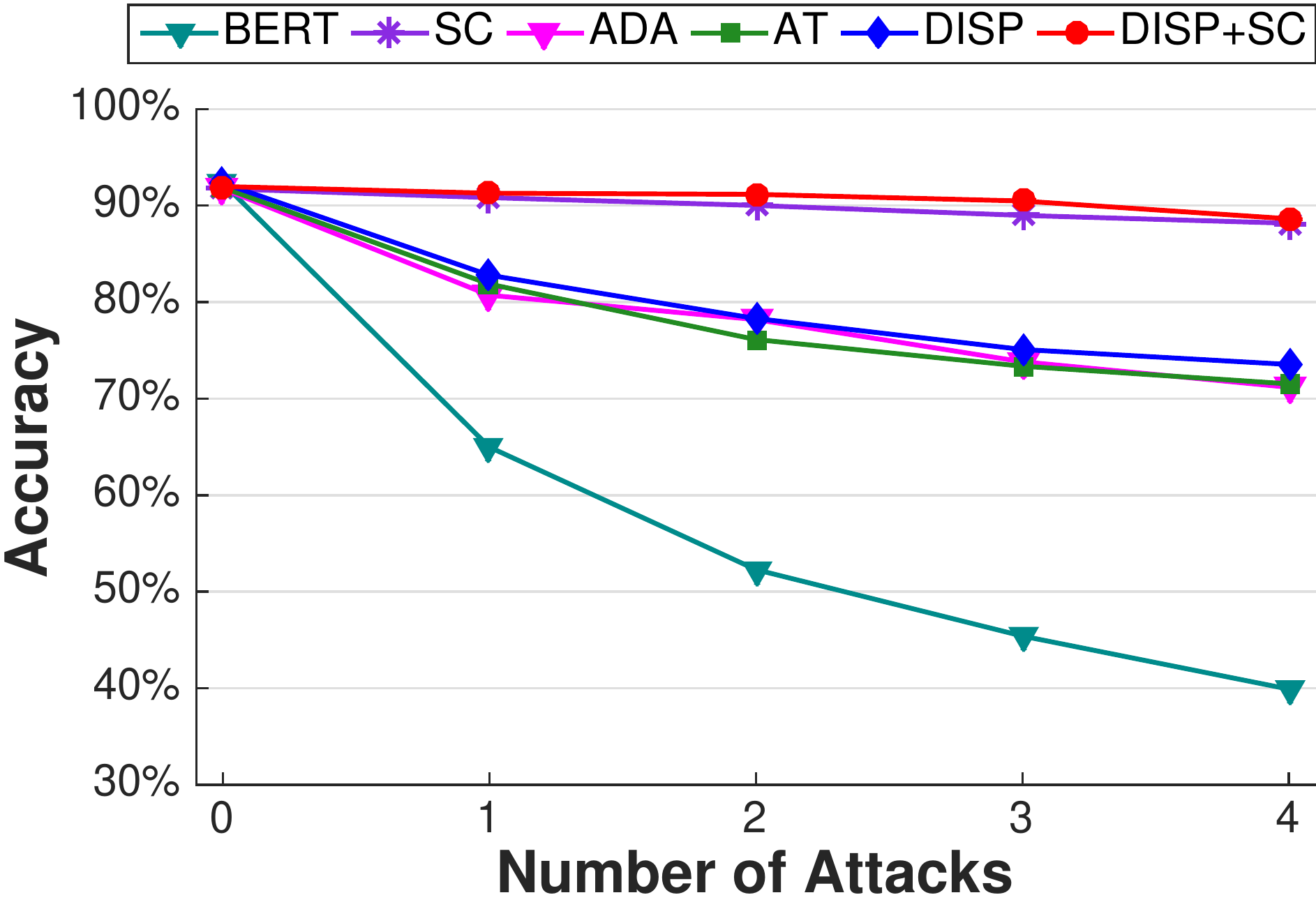}
        \caption{Insertion}
    \end{subfigure}~
    \begin{subfigure}[t]{.33\linewidth}
        \includegraphics[width=\linewidth]{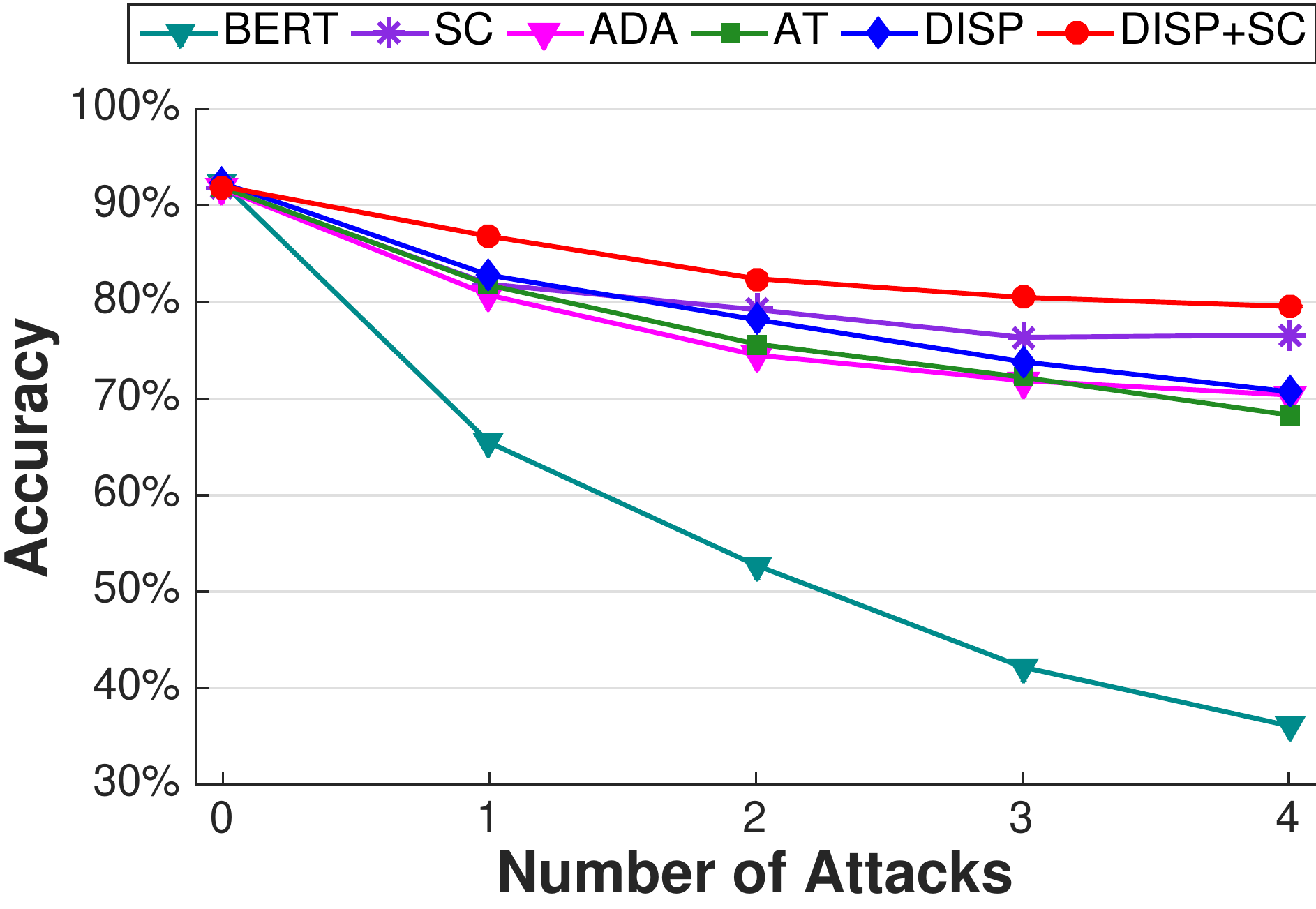}
        \caption{Deletion}
    \end{subfigure}~
    \begin{subfigure}[t]{.33\linewidth}
        \includegraphics[width=\linewidth]{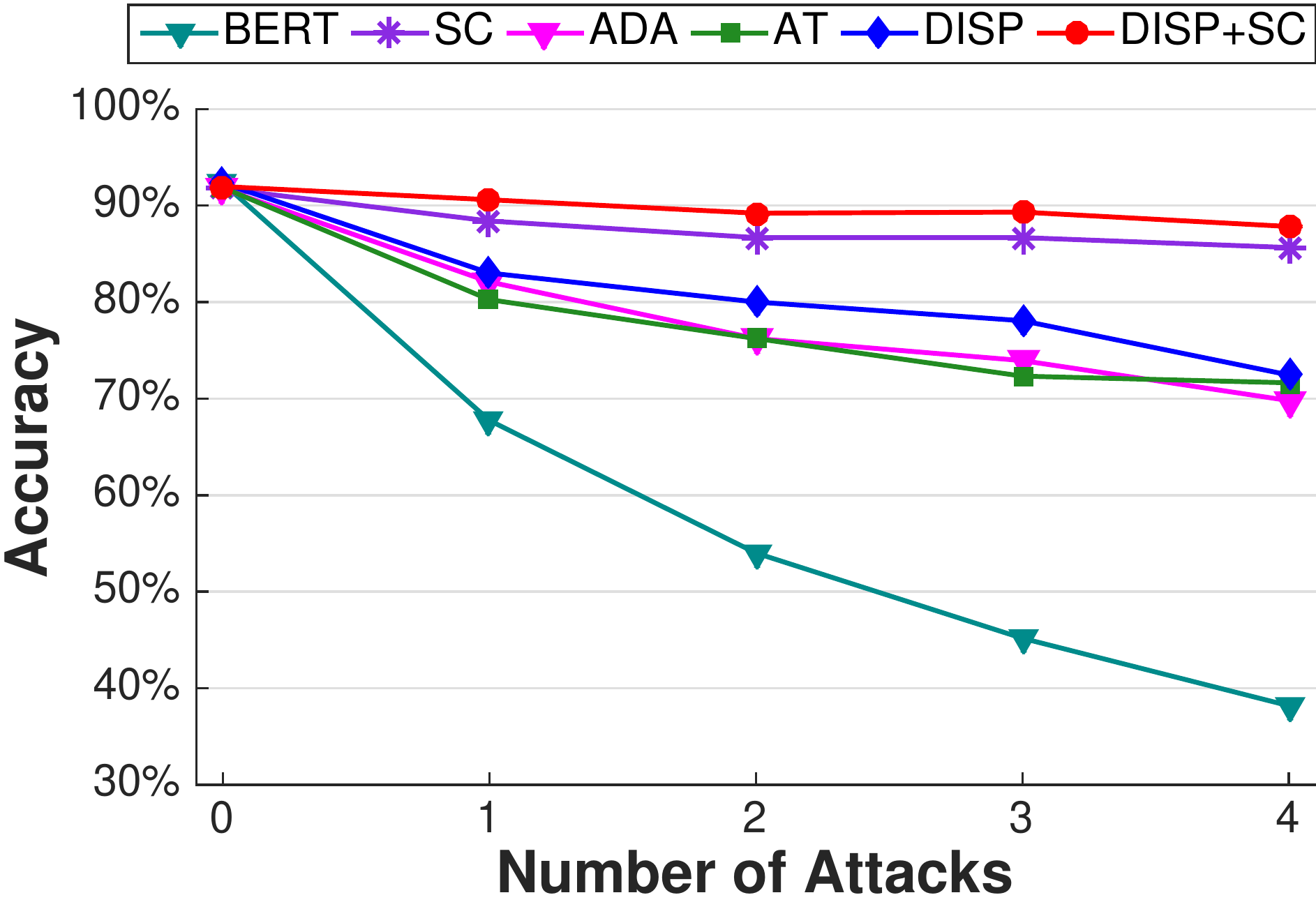}
        \caption{Swap}
    \end{subfigure}\\
        \begin{subfigure}[t]{.33\linewidth}
        \includegraphics[width=\linewidth]{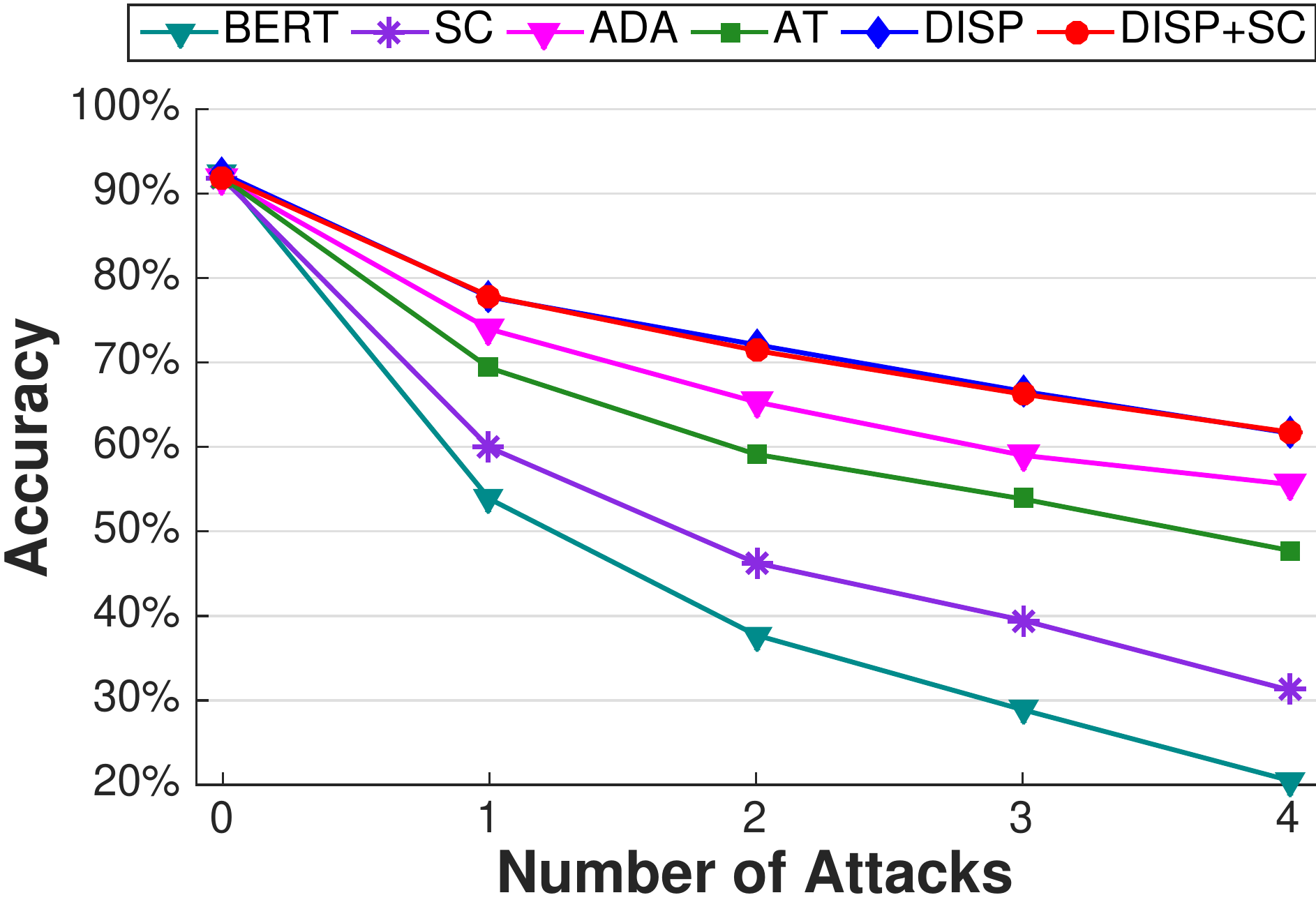}
        \caption{Random}
    \end{subfigure}~
    \begin{subfigure}[t]{.33\linewidth}
        \includegraphics[width=\linewidth]{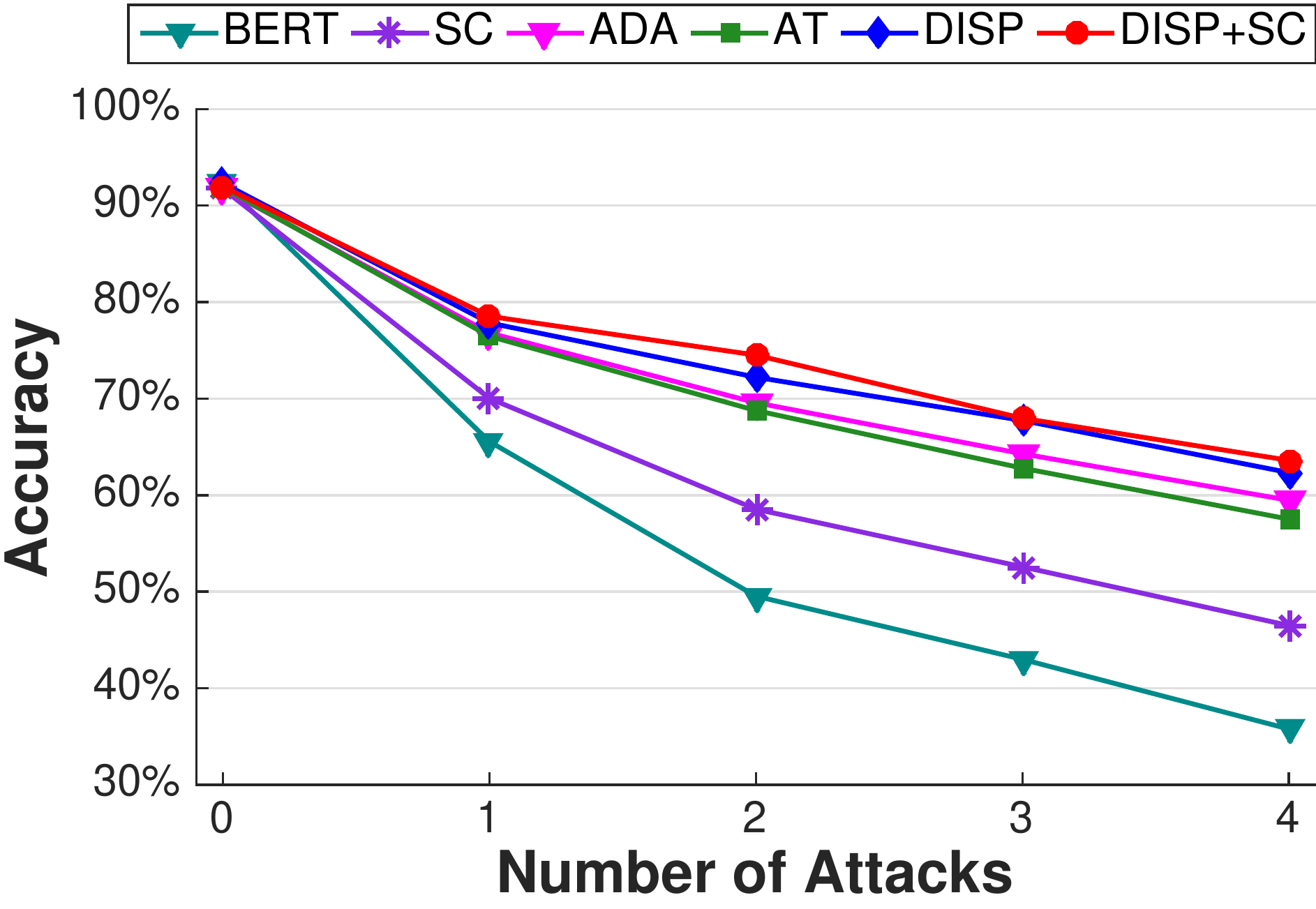}
        \caption{Embed}
    \end{subfigure}~
    \begin{subfigure}[t]{.33\linewidth}
        \includegraphics[width=\linewidth]{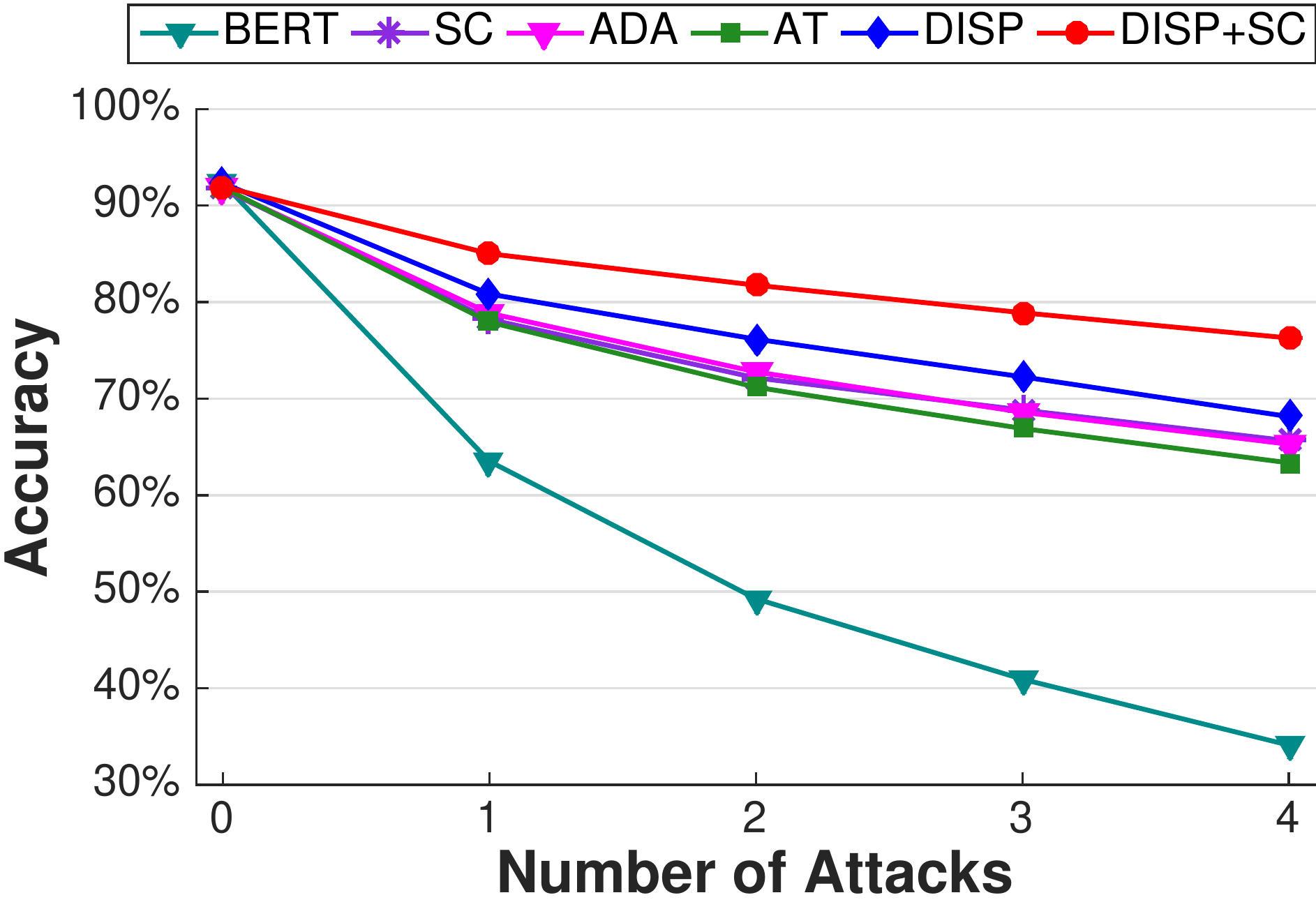}
        \caption{Overall}
    \end{subfigure}
    \caption{The accuracy of methods over different numbers and types of attacks.}
    \label{fig:attacknumber}
\end{figure*}

\input{tables/transfer_study.tex}
\input{tables/case_study.tex}

\noindent \textbf{Number of Attacks.}
Figure~\ref{fig:attacknumber} shows the classification accuracy of all methods over different numbers of attacks, i.e., perturbations, for different types of adversarial attacks.
Without using a defense method, the performance of BERT dramatically decreases when the number of attacks increases. With defense approaches, the performance drops are alleveated.  
Moreover, the relations between the performance of methods are consistent across different perturbation numbers.
DISP+SC consistently performs the best for all of the cases when DISP outperforms all of the single methods for most of the situations.
These results demonstrate the robustness of the proposed approach.

\noindent \textbf{Robust Transfer Defense.}
In practice, we may not have access to the original training corpus of a prediction model. In the following, we investigate if the perturbation discriminator can transfer across different corpora. 
We first train the discriminator and the estimator on IMDb denoted as $\text{DISP}_{\text{IMDb}}$ and then apply it to defend the prediction model on SST-2. 
Table~\ref{tab:transfer} shows the experimental results of robust transfer defense.
$\text{DISP}_{\text{IMDb}}$ achieves similar performance as the performance of $\text{DISP}_{\text{SST-2}}$ trained on the same training set. Hence, it shows that DISP can transfer the ability to recover perpetuated token across different sentiment copora. 

\noindent \textbf{Case Study of Recovered Text.}
Table~\ref{tab:case} lists four documents from SST-2 for a case study.
We successfully recovered the attacked words from ``orignal'' and ``bet'' in the cases 1 and 2 to ``imaginative'' and ``best''.
It demonstrates that embeddings generated by the embedding estimator are robust to recover the appropriate tokens and block adversarial attacks.
However, DISP performs worse when the remaining sentence is lack of informative contexts as case~3. When multiple attacks exist, the incorrect context may also lead to unsatisfactory recoveries, e.g., DISP converts ``funny'' to ``silly'' in case~4, thus flipping the prediction.
This experiment depicts a disadvantage of DISP and demonstrates that DISP+SC can gain further improvements.

\input{tables/estimator.tex}
\noindent \textbf{Embedding Estimator.}
Although DISP is not required to recover the ground-truth perturbed tokens, the embedding estimator plays an important role to derive appropriate embedding vectors that obtain the original semantics. We first evaluate the performance of embedding estimator as a regression task. The RMSE scores of estimated embeddings are 0.0442 and 0.1030 in SST-2 and IMDb datasets, which are small enough to derive satisfactory tokens.
To further demonstrate the robustness of the embedding estimator and estimated embeddings, we identify the perturbations with our discriminator and replace them with the ground-truth tokens. Table~\ref{tab:estimator} shows the accuracy scores 
over different types of attacks in the SST-2 dataset. DISP and DISP$_G$ denotes the recovery performance with our estimator and gound-truth tokens, respectively. 
More specifically, the accuracy of DISP$_G$ presents the upperbound performance gained by the embedding estimator.
The experimental results demonstrate the robustness of the embedding estimator while the estimated embeddings only slightly lower the accuracy of DISP.

\input{tables/cola.tex}

\noindent \textbf{Linguistic Acceptability Classification.}
In addition to the task of sentiment analysis, we also evaluate the performance of DISP in linguistic acceptability classification. The Corpus of Linguistic Acceptability (CoLA) is a binary classification task. The goal of this task is to predict whether an English sentence is linguistically “acceptable” or not~\cite{warstadt2018neural}.
Table~\ref{tab:cola} presents the accuracy scores of BERT and DISP on the CoLA dataset with one adversarial attack of each type.
It is interesting that the original BERT is extremely vulnerable to the adversarial attacks.
This is because the linguistic acceptability can be easily affected by perturbations.
The experimental results also depict that DISP can significantly alleviate the performance drops.
DISP is capable of blocking adversarial attacks across different NLP tasks.

%% file: tables/datasets.tex
\begin{table}[!t]
    \centering
    \resizebox{\linewidth}{!}{
    \begin{tabular}{|c|c|c|ccc|}
    \hline
        \multirow{2}{*}{Dataset} & \multirow{2}{*}{Train} & \multirow{2}{*}{Test} & \multicolumn{3}{c|}{Length} \\ \cline{4-6} 
            & & & Max. & Min. & Avg.  \\ \hline
        SST-2   &   67,349  &   1,821     &   56    &  1   &   19 \\
        IMDb    &   25,000  &   25,000  &   2,738 &  8   &   262 \\
        \hline
         
    \end{tabular}}
    \caption{The statistics of datasets.}
    \label{tab:dataset}
\end{table}

%% file: tables/attacks.tex
\begin{table}[]
    \centering
    \resizebox{\linewidth}{!}{
    \begin{tabular}{|c|c|}
    \hline
        Attack Type & Example \\
        \hline
        No Attack & Old-form moviemaking at its best. \\
        Insertion &  Old-form moviemaking at its \textbf{be{\color{red}a}st}.\\
        Deletion & Old-form moviemaking at its \textbf{be{\color{red}\sout{ s }}t}.\\
        Swap & Old-form moviemaking at its \textbf{be{\color{red}ts}}.\\
        Random & Old-form moviemaking at its \textbf{\color{red}aggrandize}.\\
        Embed & Old-form moviemaking at its \textbf{\color{red}way}.\\
        \hline
    \end{tabular}}
    \caption{Examples of each type of attack}
    \label{tab:attacks}
\end{table}

%% file: tables/discriminator_exp.tex
\begin{table*}[!t]
    \centering
    
    \resizebox{.95\linewidth}{!}{
    \begin{tabular}{|c|c|c|ccc|cc|c|} \hline
        \multirow{2}{*}{Dataset} & \multirow{2}{*}{Method} & \multirow{2}{*}{Metric}   & \multicolumn{3}{c|}{Character-level Attacks} & \multicolumn{2}{c|}{Word-level Attacks}  & Overall\\ \cline{4-8} 
         & &  & Insertion & Deletion & Swap & Random & Embed & Attacks \\ \hline\hline
    \multirow{6}{*}{SST-2} & \multirow{3}{*}{SC} 
    & Precision & 0.5087 & 0.4703 & 0.5044 & 0.1612 & 0.1484 & 0.3586 \\
    & & Recall & 0.9369 & 0.8085 & 0.9151 & 0.1732 & 0.1617 & 0.5991 \\
    & & F1  & 0.6594 & 0.5947 & 0.6504 & 0.1669 & 0.1548 & 0.4452 \\
    \cline{2-9}
    & \multirow{3}{*}{DISP} 
    & Precision  & 0.9725 & 0.9065 & 0.9552 & 0.8407 & 0.4828 & 0.8315\\
    & & Recall & 0.8865 & 0.8760 & 0.8680 & 0.6504 & 0.5515 & 0.7665\\
    & & F1 & \textbf{0.9275} & \textbf{0.8910} & \textbf{0.9095} & \textbf{0.7334} & \textbf{0.5149} & \textbf{0.7952}\\
    \hline \hline
    \multirow{6}{*}{IMDb} & \multirow{3}{*}{SC} 
    & Precision  & 0.0429 & 0.0369 & 0.0406 & 0.0084 & 0.0064 & 0.0270\\
    & & Recall & 0.9367 & 0.8052 & 0.8895 & 0.1790 & 0.1352 & 0.5891\\
    & & F1  & 0.0820 & 0.0706 & 0.0777 & 0.0161 & 0.0122 & 0.0517\\
    \cline{2-9}
    & \multirow{3}{*}{DISP} 
    & Precision & 0.9150 & 0.8181 & 0.8860 & 0.5233 & 0.2024 & 0.6690 \\
    & & Recall & 0.5068 & 0.4886 & 0.5000 & 0.3876 & 0.2063 & 0.4179 \\
    & & F1 & \textbf{0.6523} & \textbf{0.6118} & \textbf{0.6392} & \textbf{0.4454} & \textbf{0.2044} & \textbf{0.5106}\\
    \hline 
    \end{tabular}}
    \caption{Performance of SC and DISP on identifying perpetuated tokens.}
    
    \label{tab:disc_exp}
\end{table*}

%% file: tables/mainexp.tex
\begin{table*}[!t]
    \centering
    
    \resizebox{.95\linewidth}{!}{
    \begin{tabular}{|c|c|c|ccc|cc|c|} \hline
        \multirow{2}{*}{Dataset} & \multirow{2}{*}{Method} & Attack  & \multicolumn{3}{c|}{Character-level Attacks} & \multicolumn{2}{c|}{Word-level Attacks}  & Overall\\ \cline{4-8} 
         & & -free & Insertion & Deletion & Swap & Random & Embed & Attacks \\ \hline\hline
    \multirow{6}{*}{SST-2} 
    & BERT & {\bf 0.9232} & 0.6498 & 0.6544 & 0.6774 & 0.5385 & 0.6556 & 0.6351 \\ \cline{2-9}
    & SC   & 0.9174 & {\bf 0.9082} & 0.8186 & {\bf 0.8840} & 0.5993 & 0.7003 & 0.7821 \\
    & ADA   & 0.9174 & 0.8071 & 0.8071 & 0.8209 & 0.7394 & 0.7681 & 0.7885 \\ 
    & AT   & 0.9186 & 0.8186 & 0.8175 & 0.8025 & 0.6935 & 0.7646 & 0.7793\\ 
    & DISP & {\bf 0.9232} & 0.8278 & {\bf 0.8278} & 0.8301 & {\bf 0.7773} & {\bf 0.7784} & {\bf 0.8083}\\ \cline{2-9}
    & DISP+SC & 0.9197 & {\bf 0.9128} & {\bf 0.8681} & {\bf 0.9060} & {\bf 0.7784} & {\bf 0.7853} & {\bf 0.8501}\\ 
    \hline \hline
    \multirow{6}{*}{IMDb} 
    & BERT & {\bf 0.9431} & 0.8586 & 0.8599 & 0.8568 & 0.8468 & 0.8615 & 0.8567\\
    \cline{2-9}
    & SC  & 0.9193 & 0.8834 & 0.8794 & 0.8825 & 0.8695 & 0.8753 & 0.8780\\
    & ADA & { 0.9393} & 0.8766 & 0.8765 & 0.8754 & 0.8722 & 0.8755 & 0.8752\\
    & AT  & 0.8998 & 0.8958 & 0.8822 & 0.8787 & 0.8886 & 0.8822 & 0.8855\\
    & DISP & 0.9378 & {\bf 0.9310} & {\bf 0.9297} & {\bf 0.9301} & {\bf 0.9281} & {\bf 0.9347} & {\bf 0.9307}\\
    \cline{2-9}
    & DISP+SC & { 0.9395} & {\bf 0.9316} & 0.8772 & {\bf 0.9313} & 0.8755 & 0.9292 & 0.9090\\
    \hline
    \end{tabular}}
    \caption{The accuracy scores of methods with different adversarial attacks on two datasets.}
    \label{tab:main_exp}
\end{table*}

%% file: tables/transfer_study.tex
\begin{table}[!t]
    \centering
    
    \begin{tabular}{|c|c|c|c|}
    \hline
        Method & Insertion & Delete & Swap  \\
        \hline\hline
        BERT                & 0.6498 & 0.6544 & 0.6774 \\
        DISP$_{\text{SST-2}}$ &  0.8278 & 0.8278 & 0.8301 \\
        \hline
        DISP$_{\text{IMDb}}$ &	0.8243 & 0.8197 & 0.8278 \\
         \hline\hline
        Method & Random &	Embed & Overall \\
        \hline\hline
        BERT                & 0.5385 & 0.6556 & 0.6351 \\
        DISP$_{\text{SST-2}}$ & 0.7773 & 0.7784 & 0.8083 \\
        \hline
        DISP$_{\text{IMDb}}$ & 0.7623 & 0.7681 & 0.8005 \\
        \hline
    \end{tabular}
    \caption{The accuracy of DISP over different types of attacks on the SST-2 dataset with the tokens recovered by the perturbation discriminator and the embedding estimator trained on the IMDb dataset for robust transfer defense. Note that DISP$_{x}$ indicates the framework is established on the dataset $x$.}    
    \label{tab:transfer}
\end{table}

%% file: tables/case_study.tex
\begin{table*}[!t]
    \centering
    \resizebox{\linewidth}{!}{
    \begin{tabular}{|c|c|c|c|c|}
    \hline
        \# & Attacked Sentence & Recovered Token & Label & Pred \\ 
         \hline
         1 & Mr. Tsai is a very { \textbf{orig{\color{red}\sout{ i }}nal}} artist in his medium, and what time is it there?  &  imaginative  & positive    &  positive  \\
         2 & Old-form moviemaking at its \textbf{be{\color{red}\sout{ s }}t}.   &  best  & positive  &   positive \\
         3 & My reaction in a word:  \textbf{disappo{\color{red}ni}tment}.  &  that  &  negative  &  positive \\
         4 & a \textbf{painful{\color{red}i}ly} \textbf{fun{\color{red}t}ny} ode to \textbf{{\color{red}g}bad} behavior. & painfully; silly; one & positive & negative \\
         \hline
    \end{tabular}}
    \caption{A case study of recovered tokens in SST-2. Note that Label and Pred represent the ground-truth label and the predicted label.}
    \label{tab:case}
    
    \vspace{-12pt}
\end{table*}

%% file: tables/estimator.tex
\begin{table}[!t]
    \centering
    
    \begin{tabular}{|c|c|c|c|}
    \hline
        Method & Insertion & Delete & Swap  \\
        \hline
        \hline
        DISP$_G$  & 0.8773 & 0.8681 & 0.8796\\
        DISP & 0.8278 & 0.8278 & 0.8301 \\
        \hline
        Method & Random &	Embed & Overall \\
        \hline
        \hline 
        DISP$_G$ & 0.7970 & 0.7924 & 0.8429 \\
        DISP & 0.7773 & 0.7784 & 0.8083 \\
        \hline
    \end{tabular}
    \caption{The performance of DISP using ground-truth and recovered tokens over different types of attacks in SST-2. Result are in accuracy.  Note that DISP$_G$ denotes DISP using ground-truth tokens.}
    \label{tab:estimator}
\end{table}
 

%% file: tables/cola.tex
\begin{table}[!t]
    \centering
    
    \begin{tabular}{|c|c|c|c|}
    \hline
        Method & Insertion & Delete & Swap  \\
        \hline
        \hline
        BERT & 0.1160 & 0.1407 & 0.1806\\
        DISP & 0.5856 & 0.5684 & 0.6008\\
        \hline
        Method & Random &	Embed & Overall \\
        \hline
        \hline
        BERT & 0.0855 & 0.0817 & 0.1209\\
        DISP & 0.4848 & 0.5114 & 0.5502\\
        \hline
    \end{tabular}
    \caption{The accuracy scores of BERT and DISP over different types of attacks on the CoLA dataset for the task of linguistic acceptability classification. The accuracy score of BERT without any attack is 0.8519.}    
    \label{tab:cola}
\end{table}

%% file: sections/s5-conclusions.tex
\section{Conclusions}
\label{section:conclusions}

In this paper, we propose a novel approach to discriminate perturbations and recover the text semantics, thereby blocking adversarial attacks in NLP.
DISP not only correctly identifies the perturbations but also significantly alleviates the performance drops caused by attacks.